\documentclass[10pt,conference]{IEEEtran}

\IEEEoverridecommandlockouts
\usepackage[T1]{fontenc}
\usepackage{cite}
\usepackage{amsmath,amssymb,amsfonts}
\usepackage{algpseudocode}
\usepackage{algorithm}
\usepackage{graphicx}
\usepackage{textcomp}
\usepackage{xcolor}
\usepackage{url}
\usepackage{subcaption}
\usepackage{listings,xcolor}
\usepackage{adjustbox}
\usepackage{booktabs}  
\usepackage{array}     
\usepackage{longtable} 
\usepackage{soul}

\usepackage[numbered]{matlab-prettifier}

\usepackage[colorlinks=true, linkcolor=blue, citecolor=blue, urlcolor=blue]{hyperref}

\colorlet{punct}{red!60!black}
\definecolor{background}{HTML}{EEEEEE}
\definecolor{delim}{RGB}{20,105,176}
\colorlet{numb}{magenta!60!black}

\lstdefinelanguage{JSON}{
    basicstyle=\tiny\ttfamily,
    numbers=left,
    numberstyle=\scriptsize,
    stepnumber=1,
    numbersep=8pt,
    showstringspaces=false,
    string=[s]{"}{"},
    breaklines=true,
    frame=lines,
    backgroundcolor=\color{background},
    string=[s]{"}{\"},
    literate=
     *{0}{{{\color{numb}0}}}{1}
      {1}{{{\color{numb}1}}}{1}
      {2}{{{\color{numb}2}}}{1}
      {3}{{{\color{numb}3}}}{1}
      {4}{{{\color{numb}4}}}{1}
      {5}{{{\color{numb}5}}}{1}
      {6}{{{\color{numb}6}}}{1}
      {7}{{{\color{numb}7}}}{1}
      {8}{{{\color{numb}8}}}{1}
      {9}{{{\color{numb}9}}}{1}
      {:}{{{\color{punct}{:}}}}{1}
      {,}{{{\color{punct}{,}}}}{1}
      {\{}{{{\color{delim}{\{}}}}{1}
      {\}}{{{\color{delim}{\}}}}}{1}
      {[}{{{\color{delim}{[}}}}{1}
      {]}{{{\color{delim}{]}}}}{1},
}

\def\BibTeX{{\rm B\kern-.05em{\sc i\kern-.025em b}\kern-.08em
    T\kern-.1667em\lower.7ex\hbox{E}\kern-.125emX}}

\begin{document}

\title{
Why do AI agents communicate in human language?
}

\author{
\IEEEauthorblockN{Pengcheng Zhou, Yinglun Feng, Halimulati Julaiti, Zhongliang Yang\textsuperscript{*}\thanks{\textsuperscript{*}Corresponding author.}}
\IEEEauthorblockA{\textit{Beijing University Of Posts and Telecommunications}}
}

\maketitle
\begin{abstract}
Large Language Models (LLMs) have become foundational to modern AI agent systems, enabling autonomous agents to reason and plan. In most existing systems, inter-agent communication relies primarily on natural language. While this design supports interpretability and human oversight, we argue that it introduces fundamental limitations in agent-to-agent coordination. The semantic space of natural language is structurally misaligned with the high-dimensional vector spaces in which LLMs operate, resulting in information loss and behavioral drift. Beyond surface-level inefficiencies, we highlight a deeper architectural limitation: current LLMs were not trained with the objective of supporting agentic behavior. As such, they lack mechanisms for modeling role continuity, task boundaries, and multi-agent dependencies. The standard next-token prediction paradigm fails to support the structural alignment required for robust, scalable agent coordination. Based on this, we argue that two core questions deserve careful examination: first, given that AI agents fundamentally operate in high-dimensional vector spaces, should they rely on a language system originally designed for human cognition as their communication medium? Second, should we consider developing a new model construction paradigm that builds models from the ground up to natively support structured communication, shared intentionality, and task alignment in multi-role, multi-agent environments? This paper calls for a reconsideration not only of how agents should communicate, but also of what it fundamentally means to train a model that natively supports multi-agent coordination and communication.
\end{abstract}


\section{Introduction \label{sec:introduction}}




In recent years, artificial intelligence agent systems have demonstrated significant impact across a wide range of real-world applications, including collaborative software development~\cite{wasif2025multi}, virtual assistants~\cite{financialtimes2025aiagents}, multi-robot coordination~\cite{li2025large}, decision support systems~\cite{relevanceai2024mas}, and agent control in gaming environments~\cite{wikipedia_multiagent}. With the rapid advancement of Large Language Models (LLMs), these models have been extensively integrated into agent systems~\cite{lyu2025llmcoop} as core modules for reasoning and decision-making, significantly enhancing agents' capabilities in planning, inference, and response under complex task settings.

In multi-agent systems, the design of communication mechanisms is key to achieving effective coordination. To this end, a substantial body of research has focused on developing multi-agent communication mechanisms, including Multi-agent Communication Protocols (MCPs) and Agent-to-Agent (A2A) communication frameworks~\cite{ehtesham2025survey,google2025a2a,anthropic2024mcp}. These approaches aim to enable agents to share information, divide tasks, and achieve joint goals in asynchronous environments. A prevailing practice in current systems is to employ natural language as the primary medium for agent communication~\cite{yi2024survey}. Leveraging the language understanding and generation capabilities of LLMs, many systems adopt multi-turn, dialogue-like interaction patterns~\cite{autogen2025}. This strategy simplifies system interface design, enhances human readability, and facilitates human-in-the-loop supervision and intervention~\cite{zhou2025continuum}.

However, as system complexity and task scale increase, an emerging body of research and empirical evidence reveals significant limitations in agent systems that rely on natural language communication~\cite{guan2025evaluating, laban2025llms,han2024llm}. For example, systems such as AutoGPT and AgentVerse frequently exhibit issues such as goal drift, broken task logic, redundant or ambiguous language output, and incoherent multi-step planning~\cite{pungas2023autogpt,agentverse2024}. Several studies have pointed out that natural language protocols struggle to accurately represent complex task semantics, often leading to inconsistencies between what agents say and what they actually do~\cite{cemri2025multi,zhang2025agent,turpin2023language,chen2025reasoning,wang2023fake}. Other works have shown that natural language introduces substantial communication overhead, negatively impacting system responsiveness and resource efficiency~\cite{zhang2024cut,galileo2025multiagent}.

This paper reconsiders the rationale of using natural language as the communication protocol among agents, focusing on a central question: should artificial agents, which fundamentally operate in high-dimensional vector spaces, rely on a communication system originally designed for human cognitive needs? We argue that the representational space of natural language is structurally misaligned with the internal semantic space of LLMs~\cite{mordatch2018emergence,futrell2022information}. This misalignment can lead to semantic compression, information loss, and failure in maintaining state consistency during communication, thereby constraining task coordination and semantic sharing in multi-agent systems~\cite{lazaridou2020multi,hu2025position,yan2025beyond}.

Furthermore, we suggest that this issue reflects a deeper limitation inherent in the architecture of current LLMs. These models were never trained to function as native agents in multi-role interactive environments. Their training paradigm, which is based on next-token prediction over linear textual streams, lacks support for modeling task boundaries, maintaining role continuity, and capturing inter-agent dependency structures. Based on this, we raise an extended question: should we explore a new model training paradigm that, from the ground up, incorporates structured communication mechanisms, shared intentionality modeling, and task alignment objectives across agents, in order to train AI systems with native capabilities for collaboration? This paper aims to initiate a systematic rethinking of both agent communication mechanisms and model design paradigms, providing a foundation for the development of next-generation multi-agent systems. To the best of our knowledge, this is the first systematic analysis connecting the structural mismatch between LLMs’ training paradigms and the requirements of scalable agent coordination and highlighting the need for a new modeling paradigm that supports semantic alignment and role persistence in multi-agent systems.

\section{Related Work}

\subsection{LLM-based Agent Systems}
In recent years, with the rapid development of Large Language Models (LLMs), an increasing number of multi-agent systems (MAS) have been built upon LLMs. These systems typically leverage the powerful language understanding and generation capabilities of LLMs to enable inter-agent reasoning, task planning, and role coordination. Representative general-purpose frameworks include AutoGen\cite{autogen2025}, AgentVerse\cite{agentverse2024}, CAMEL\cite{li2023camel}, and CrewAI\cite{crewai2024}. These systems support multi-turn natural language interactions among agents and have been applied in various collaborative tasks such as code generation, data analysis, and open-domain question answering.

Beyond general frameworks, several domain-specific multi-agent systems have emerged. For example, in the financial domain, TradingAgents\cite{xiao2024tradingagents} introduces a collaborative trading framework where LLM-powered agents specialize in fundamental, sentiment, and technical analysis to enhance portfolio performance. In the domain of mental health support, a multi-agent dual dialogue system\cite{kampman2024ai} enables AI agents to assist therapists through empathetic response suggestion and contextual analysis. In education, SimClass~\cite{zhang2024simulating} simulates role-based classroom interactions among teacher, assistant, and student agents, providing structured control over instructional dynamics in intelligent tutoring environments.

These systems demonstrate the considerable potential of LLMs in multi-agent modeling. However, despite their increasingly sophisticated functionalities, they share a consistent architectural design: natural language is used as the default and primary medium of communication among agents. While this design is intuitively reasonable in early-stage systems—given natural language's advantages in interpretability, debugging, and human-in-the-loop integration—this communication paradigm has shown significant limitations as system complexity and task scale increase. Common issues include semantic redundancy, ambiguity in interpretation, lossy information compression, and difficulty maintaining inter-agent state consistency~\cite{cemri2025multi, han2024llm, laban2025llms, turpin2023language,chen2025reasoning}. Current research largely focuses on incremental improvements to existing systems, such as optimizing prompt engineering~\cite{wei2022chain} or integrating external memory modules~\cite{mialon2023augmented,shinn2023reflexion} to enhance communication effectiveness.  While these efforts improve system stability to some extent, they remain patch-based solutions that do not address the underlying representational misalignment between LLMs and agent communication needs. Critically, the question of whether natural language is fundamentally suitable as an inter-agent communication protocol remains largely underexplored. In particular, few studies have systematically explored whether current LLM architectures are structurally aligned with the representational requirements of scalable multi-agent coordination. This gap suggests a need for more foundational investigation into how communication mechanisms and model design paradigms co-evolve in agentic settings.

\subsection{Communication in Multi-Agent Systems}
Communication mechanisms have long been a central component in the design of Multi-Agent Systems (MAS), serving as the foundation for coordination and autonomy among agents. Early MAS architectures typically employed structured, symbolic communication protocols to facilitate inter-agent interaction and cooperation. Representative examples include the Agent Communication Language (ACL) and Knowledge Query and Manipulation Language (KQML), which define precise semantic frameworks, preconditions, and expected effects for each message type~\cite{finin1994kqml, fipa2000fipa}. These protocols ensure that messages are interpretable, executable, and semantically consistent within the system. In such frameworks, communication itself is tightly coupled with task logic, agent states, and behavioral plans~\cite{wooldridge2009introduction}. In contrast to symbolic protocols, recent advances in deep multi-agent reinforcement learning (MARL) have explored the emergence of communication strategies directly from agent interaction, allowing agents to learn message protocols in an end-to-end differentiable fashion~\cite{foerster2016learning}. These emergent systems prioritize minimal signaling necessary for successful coordination, and are explicitly optimized for internal state alignment and task completion—rather than human interpretability or linguistic fluency. 

In the domain of multi-agent reinforcement learning and collective intelligence, researchers have also explored emergent communication~\cite{lazaridou2020emergent, lowe2019pitfalls, foerster2016learning}, where agents learn to generate discrete signals through training in task-specific environments. These emergent protocols emphasize minimal redundancy and maximal task-specific coupling. At their core, they reflect a foundational principle: communication systems designed for machines should serve the execution of tasks and the synchronization of internal states—rather than presentation for human interpretation.

However, current AI agents built upon LLMs are gradually deviating from this design philosophy. In recent years, LLM-based agents have increasingly relied on natural language as the dominant medium of communication, particularly for multi-turn coordination, planning, and information exchange. To organize this natural language interaction process, a number of frameworks have introduced standardized communication protocols tailored for AI agents, such as the Multi-Agent Communication Protocol (MCP)\cite{anthropic2024mcp} and the Agent-to-Agent Communication Framework (A2A)\cite{google2025a2a}. These protocols aim to encapsulate language-based interactions through structured message formats, dialogue turns, and callable function semantics, thereby enabling smoother coordination among LLM-powered agents.

It is important to note, however, that natural language is not inherently designed for machine-level interaction. Its semantics are intrinsically ambiguous, polysemous, and redundant. Compared to structured symbolic protocols, natural language communication is more prone to semantic drift, desynchronized agent states, and inconsistent behaviors. These issues become particularly pronounced in scenarios involving multiple roles and complex task dependencies~\cite{lazaridou2020multi,kottur2017natural,jacob2021multitasking,smythos2025comparing}. When agents rely on natural language to convey plans, intentions, or internal states, they face substantial degradation in communication efficiency, interpretability precision, and behavioral fidelity. In such settings, the system may exhibit the superficial appearance of dialogue, while in reality, coordination is weakened or even lost. Although protocols like MCP and A2A provide structural scaffolding for natural language interaction, they remain fundamentally dependent on human-oriented language as the underlying carrier. In tightly coupled environments with high-frequency interactions, natural language protocols often result in increased redundancy, degraded synchronization, and reduced execution alignment~\cite{piatti2024cooperate,zou2025survey,brown2017multiagent}. Recent work further highlights that even seemingly similar large language models (e.g., GPT-3.5 vs. GPT-4) exhibit idiosyncratic semantic boundaries, manifesting divergent behaviors under aligned prompts and tasks~\cite{sun2025idiosyncrasies}. Such divergence underscores the difficulty of establishing shared intent or synchronized understanding in multi-agent settings when relying solely on natural language. These findings reinforce the argument that natural language, while expressive for human communication, introduces fragility and inconsistency when used as the coordination substrate for artificial agents.

We argue, therefore, that communication among AI agents should not be centered solely around mimicking human conversation. Rather, it should return to the core functional objective of agent communication: the efficient, reliable, and executable transmission of state and intent between machines. While contemporary natural language-based protocols such as MCP and A2A have made meaningful engineering progress, they have not yet resolved key challenges related to semantic fidelity and state alignment. At its root, this limitation may not lie in the protocol interface itself, but rather in the underlying model architecture, which is not inherently designed to support structured coordination across autonomous agents.

\subsection{Limitations of Natural Language in Agent Coordination}



In current multi-agent systems, natural language is widely adopted as the primary medium of communication among AI agents. However, as system complexity increases and the demand for deep task-level coordination intensifies, several critical limitations of natural language in agent collaboration are becoming increasingly evident. The root of the issue lies in the fact that natural language was fundamentally designed for human communication. It prioritizes ambiguity tolerance, emotional nuance, and pragmatic flexibility—features well-suited for informal and socially embedded human interactions. In contrast, agent-to-agent coordination relies on precise state transmission, role persistence, behavioral consistency, and explicit intent alignment. These fundamentally different communication goals introduce structural tensions between the affordances of natural language and the requirements of artificial agents. Recent studies have highlighted these limitations from various dimensions.

First, the unstructured nature of natural language increases the risk of information loss during multi-turn interactions. Laban et al.~\cite{laban2025llms} describe the "lost-in-conversation" phenomenon, where agents gradually lose track of context and task chains during dialogue-based planning, leading to broken logical flows. This issue is particularly prominent in systems like AutoGen.

Second, natural language is not inherently bound to environment states or executable actions, often resulting in a disconnect between language and behavior. Cemri et al.~\cite{cemri2025multi} found that LLM-based agents frequently "describe" the successful completion of a task without actually executing the required action or updating the environment state, revealing a gap between language generation and actual agent behavior.

Third, the redundancy and verbosity of natural language can interfere with reasoning efficiency. Birr et al. ~\cite{birr2024autogpt+} observed that in the AutoGPT+P system, agents often enter reasoning loops during task decomposition and path planning due to repetitive confirmations, explanations, and speculative dialogue, ultimately failing to complete the assigned task. This highlights the non-goal-directed interference introduced by language. Furthermore, Microsoft Research ~\cite{laban2025llms} emphasized that the lack of explicit state modeling in natural language makes it difficult for agents to maintain coherent goal chains and context alignment in multi-step tasks, often leading to execution collapse or erratic planning jumps—especially when multiple agents are required to coordinate roles and intentions.

Finally, Han et al.~\cite{han2024llm} demonstrated that in systems involving multiple roles and task delegation, the use of natural language as the primary communication channel can lead to semantic drift and role confusion, making it difficult for agents to maintain consistent collaborative structures over time. This suggests that natural language, as currently used, lacks the capacity to support semantic alignment and role constraints required for persistent, multi-agent collaboration.

Taken together, these findings reveal that natural language imposes several systemic limitations on inter-agent communication. It not only constrains coordination efficiency, but also reflects a deeper disconnect between current communication mechanisms and underlying model architectures. These challenges cannot be resolved simply through prompt engineering or additional dialogue rounds; rather, they expose the lack of native state control and structured intent modeling capabilities in today’s language models. As such, the question we raise—should AI agents continue to communicate in human language—is not only of theoretical interest, but also of urgent practical significance.

\section{Theoretical Analysis}
\subsection{Semantic Misalignment Analysis}
In multi-agent systems, communication fundamentally entails the precise mapping and updating of informational states across agents. This process presupposes that the symbolic system used for communication is structurally isomorphic to the internal representational mechanisms of the agents. In current mainstream large language model (LLM)-powered agent systems, communication predominantly relies on natural language as the medium of information exchange. However, natural language and the internal representational space of LLMs diverge significantly in structural assumptions and semantic construction~\cite{harnad1990symbol,baroni2020linguistic}, resulting in coordination failures such as semantic compression, irreversible state mapping, and intention drift.

Natural language is a representational system based on discrete token sequences, with its elementary units defined by a finite vocabulary and realized through sparse symbolic projections during training and generation. Its semantic construction depends on human-level cognitive consensus, emphasizing pragmatic flexibility, contextual tolerance, and polysemy. Consequently, natural language as a discrete symbolic system induces sparse and non-differentiable transitions in its surface form, which stands in contrast to the smooth structure of vector manifolds built by continuous embeddings in LLMs. The internal representations of LLMs, by comparison, are not discrete linguistic symbols but rather multi-layered embeddings defined in high-dimensional, continuous, dense tensor spaces. These representations encode task states, environmental cues, action intentions, and memory, and are essentially a process of semantic tensor evolution over dynamic multimodal distributions—characterized by fluidity, differentiability, and hierarchy.

Under this semantic system mismatch, LLMs generate natural language for communication through a nonlinear projection from the high-dimensional semantic tensor space $\mathcal{H}$ to a low-dimensional discrete token space $\mathcal{L}$:

\begin{equation}
f: \mathcal{H} \rightarrow \mathcal{L}, \quad \text{where } \mathcal{L} \subset \mathbb{T}^n \text{ is a discrete token space}.
\end{equation}

This projection inevitably induces information compression, whereby multi-dimensional semantic states embedded in the original tensor representation are partially lost when mapped to natural language surface forms. As $f$ is typically many-to-one, distinct internal semantic states may correspond to identical linguistic expressions, leading to semantic aliasing and non-invertibility.

Moreover, natural language exhibits semantic drift in its contextual construction, i.e., semantic instability over multi-turn dialogues. Thus, even if agents initially share a semantic representation, continued interaction via natural language may progressively deviate from the original consensus, resulting in intention misalignment. This "semantic drift chain" often manifests in task execution as path jumping, role misjudgment, and collaboration breakdown. Specifically, let agent A have an internal state $h_A \in \mathcal{H}$ and send a natural language message $l = f(h_A)$ to agent B. Agent B encodes this message as $\hat{h}_B = f^{-1}(l)$. Given the non-invertibility of $f$, it is almost always the case that $\hat{h}_B \neq h_A$. This representational error accumulates over multiple interaction rounds, ultimately leading to behavioral divergence.

It is important to emphasize that this semantic misalignment is not merely a communication-level encoding issue but a structural tension intrinsic to the model’s generative mechanism. Existing LLMs are pre-trained with objectives that maximize the next-token conditional probability over linear sequences, rather than optimizing for state-aligned tensors or multi-agent intention graphs. This training paradigm inherently lacks structural support for role persistence, state tracking, and semantic synchronization in agent-to-agent scenarios. Even the incorporation of chain-of-thought, function-calling, or memory buffer mechanisms cannot circumvent the irreversibility and ambiguity of the $\mathcal{H} \rightarrow \mathcal{L}$ mapping.

Therefore, we argue that the fundamental limitation of natural language as a communication medium in multi-agent systems lies not in its expressive capacity, but in its inherent unsuitability as a precise alignment interface between high-dimensional internal state spaces. Without jointly modeling the semantic structure of communication and the representational architecture of agents, any system relying on natural language for internal state transfer will inevitably face cumulative coordination errors and behavioral drift risks.

Further, consider the agent communication process as a chain of continuous state transitions, where each interaction step can be modeled as:

\begin{equation}
h_{t+1}^{(i)} = \mathcal{T}(f^{-1}(f(h_t^{(j)})))
\end{equation}

Note that $f^{-1}$ here denotes an approximate decoding process, as $f$ is typically many-to-one and non-invertible in a strict mathematical sense.
Here, $h_t^{(j)}$ denotes the internal semantic state of agent $j$ at time $t$, and $\mathcal{T}$ is the state update operator. Due to the non-invertibility and compression nature of $f$, the error term $\delta_t = \| h_t^{(j)} - f^{-1}(f(h_t^{(j)})) \|$ accumulates over time. This leads to a system-level cascading semantic loss, whose total error upper bound can be represented as:

\begin{equation}
L_{\text{cascade}} = \sum_{t=1}^{T} \mathbb{E}_{j \in \mathcal{A}} \left[\| h_t^{(j)} - \hat{h}_t^{(j)} \|^2\right]
\end{equation}

where $\hat{h}_t^{(j)}$ denotes the estimated state of agent $j$ recovered from the natural language message, and $\mathcal{A}$ denotes the set of agents. Given the inevitability of semantic aliasing and linguistic ambiguity in current LLM-based communication systems, this loss admits a non-zero lower bound and cannot be eliminated under standard decoding assumptions. That is, $L_{\text{cascade}}$ cannot theoretically converge to zero without introducing additional structural priors or invertible semantic interfaces.

The structural origin of this loss lies in the mismatch of semantic spaces, not in variations of linguistic output quality~\cite{kottur2017natural, han2024llm}. Therefore, conventional methods such as prompt tuning, memory injection, or context tracking cannot fundamentally eliminate $L_{\text{cascade}}$. This underscores our central claim: 

\textbf{To mitigate cascading semantic loss, communication mechanisms must maintain structural consistency with internal representational spaces at the semantic modeling level.}

\subsection{Protocol-Induced Agent Misbehavior}
The use of natural language as a communication protocol in multi-agent systems is primarily motivated by goals of interpretability and flexibility. However, from a system behavior modeling perspective, the language protocol is not a neutral medium. Its inherent semantic indeterminacy and structural opacity can induce a range of agent behavior deviations during task execution. These deviations are not merely reflected in task failure outcomes but are rooted in mismatches within the structural mechanisms underlying agent trajectories. We define this phenomenon as \textbf{Protocol-Induced Agent Misbehavior}, which fundamentally arises from the instability of behavior mappings and the semantic vagueness of natural language execution within the system.

In LLM-based agent architectures, action decision-making typically follows a reasoning chain of semantic parsing, state assessment, and action selection based on natural language goals. Due to the openness of language descriptions and the ambiguity of internal LLM representations, two principal types of behavioral mismatches frequently occur:

\begin{itemize}
\item \textbf{(1) Goal Misinterpretation}: Natural language goals often lack formalized task constraints (e.g., "optimize efficiency") and are semantically open-ended. This allows LLMs to activate multiple inconsistent subgoal representations during internal parsing, resulting in execution path deviation and semantic drift. In multi-turn decomposition in AutoGPT, agents often hallucinate nonexistent subtasks, causing recursive planning loops or dangling task states (e.g., "write a non-existent report to analyze user intent")~\cite{birr2024autogpt+,zhang2024language,agenticAIguide}.

\item \textbf{(2) Action-State Decoupling}: Under natural language protocols, agents rely on linguistic parsing to guide action selection. However, because language tokens do not directly map to the environment state space, there exists a systematic decoupling interval between perceived state and actual behavior. Even if language expresses a "completed" or "in-progress" state, the system may not have updated its internal state or triggered environmental changes. As shown in the study by Cemri et al.~\cite{cemri2025multi}, LLM-agents often demonstrate pseudo-execution, where the language reports completion while no real system execution occurs, highlighting severe state consistency issues in language-driven behaviors.
\end{itemize}

Furthermore, protocol-induced misbehavior is cumulative and transferable. In multi-agent systems, a behavioral error induced in one agent by the language protocol can propagate to others via linguistic transmission. For instance, if agent A misinterprets a task and executes incorrectly, its generated language encodes this error as a "relay state." Agent B, without direct environment observation, continues planning based on this faulty linguistic input, forming a \textbf{misaligned coordination chain}. This is particularly prominent in systems with role-specific division of labor and interdependent subtasks.

Formally, let the decision function of an agent $i$ in the agent set $\mathcal{A}$ be defined as:

\begin{equation}
l_i = f(h_i), \quad r_i = \mathcal{R}(l_i), \quad \pi_i = \phi(r_i),
\end{equation}

where $h_i$ denotes the internal state representation of agent $i$, $f$ is the natural language projection function, $\mathcal{R}$ represents the intermediate reasoning or planning module (e.g., AutoGPT's Planner), and $\phi$ denotes the final action policy generator. Once $l_i$ (the linguistic message) misrepresents the task goal or current status, it may induce downstream errors in agent $j$.

Assuming agent $j$ updates its internal state based on $l_i$, the propagation of misaligned semantics can be modeled as:

\begin{equation}
h_j \leftarrow \mathcal{T}_j(\tilde{f}^{-1}(l_i)) \quad \Rightarrow \quad \pi_j = \phi(\mathcal{R}(f(h_j))),
\end{equation}

where $\tilde{f}^{-1}$ denotes an approximate inverse of the language projection (as $f$ is typically lossy and non-invertible), and $\mathcal{T}_j$ is the state update operator of agent $j$.

Given that $f$, $\mathcal{R}$, and $\phi$ are all nonlinear and generally lack invertibility or continuity, the system accumulates semantic approximation error over communication steps, resulting in unpredictable task trajectory deviation and plan instability.

Even in single-agent systems, execution inconsistency induced by language protocols is not a sporadic anomaly but a form of structural noise. For instance, AutoGPT frequently exhibits behavior chains such as: "generate plan $\rightarrow$ interpret subtask $\rightarrow$ produce completion feedback $\rightarrow$ no actual task execution." Though such trajectories appear coherent linguistically, analysis of environmental state feedback and execution traces reveals no closed-loop task completion~\cite{pungas2023autogpt,autogpt2023issue5190}. This suggests that the language protocol lacks effective task state supervision capability.

Therefore, we argue that natural language, when used as a communication protocol, inherently introduces uncontrollable reasoning deviations and behavioral errors. The issue extends beyond prompt engineering and stems from a structural misalignment between the language protocol and the agent system’s planning function. This architectural imbalance causes the system to drift from stable policy spaces and undermines the core requirements of reliability and controllability.

\subsection{Architectural Incompatibility with Multi-Agent Requirements}
Coordinated behavior in multi-agent systems depends not only on the expressive capacity of the communication protocol, but more critically on the structural capabilities of individual agents to maintain stable role identities, decompose and allocate tasks, and align behavioral intentions throughout interaction. However, current systems based on large language models (LLMs) are fundamentally misaligned with these requirements at both the modeling paradigm and representational level. The single-channel, single-perspective architecture of LLMs introduces a structural mismatch when applied to collaborative scenarios that inherently involve multiple roles, distributed states, and dynamic inter-agent dependencies.

From a training perspective, standard LLM architectures are optimized under the objective of maximizing the conditional likelihood of tokens in a linear sequence. Formally, the learning objective is defined as:

\begin{equation}
\mathcal{L}_{\text{LM}} = -\sum_{t=1}^{T} \log P(x_t \mid x_{<t}; \theta),
\end{equation}

where $x_{1:T}$ denotes the input token sequence, and $\theta$ represents the model parameters. This formulation presumes a uni-directional, context-free stream of tokens and lacks structural priors necessary for modeling role identities, inter-agent dependencies, or multi-threaded state transitions. This objective is defined over unstructured token streams and does not differentiate between semantic senders and receivers. Consequently, in multi-agent contexts, the model cannot maintain a consistent binding between an agent's internal state $h_t^{(i)}$ and its role identifier $r^{(i)}$, that is:

\[
\neg \exists f : r^{(i)} \rightarrow h_t^{(i)} \quad \text{s.t. } f(r^{(i)}) \approx h_t^{(i)} \text{ over } t.
\]

As a consequence, a single model may exhibit role drift and behavioral attribution failures across multi-turn interactions. This phenomenon is particularly pronounced in LLM-based dialogue systems, where role confusion and instructional boundary violations frequently occur—even when system prompts explicitly delineate the responsibilities and identities of different agents. Such behavior suggests that the model’s internal representation space $\mathcal{H}$ lacks structural mechanisms for semantic boundary preservation.

Moreover, task decomposition in multi-agent settings is not merely a matter of instruction parsing. It presupposes an internal representational space that supports decomposability and local operability. Let the global task state be denoted by $S$. Ideally, it should admit a principled decomposition into a set of agent-specific subspaces $\{S^{(i)}\}_{i=1}^n$ such that:

$$
S = \bigoplus_{i=1}^n S^{(i)}, \quad \text{where } \mathcal{T}(S^{(i)}) \subseteq S^{(i)},
$$

where $\mathcal{T}$ denotes the local transition dynamics. However, the latent representation space in LLM-based architectures is defined as a unified high-dimensional continuous tensor space $\mathcal{H} \subset \mathbb{R}^d$, which lacks structural inductive biases required for subgraph segmentation or task dependency modeling. This architectural design offers no native support for partitioning semantic goals or for encoding operational boundaries between agents. As a result, during interaction, multiple agents may redundantly generate actions toward the same subgoal or take conflicting execution paths—symptomatic of the system’s inability to enforce structurally grounded task decomposition.

Furthermore, action policies in LLM-based agents are independently instantiated. That is, the policy of agent $i$, denoted $\pi_i(a \mid s)$, is solely a function of its private interaction history $x_{1:T}^{(i)}$ and shared model parameters $\theta$. There exists no internal mechanism for inter-agent policy synchronization or alignment. Let $\Delta_{\text{intent}}^{(i,j)}$ quantify the semantic divergence between agent $i$ and agent $j$'s intended actions under the same state $s$:

$$
\Delta_{\text{intent}}^{(i,j)} = \mathbb{E}_{s \sim \mathcal{S}} \left[\| \pi_i(a|s) - \pi_j(a|s) \|^2 \right],
$$

In LLM-based agents, each policy $\pi_i(a \mid s)$ is independently constructed from the agent’s past sequence $x_{1:T}^{(i)}$ using shared parameters $\theta$, with no shared latent graph or alignment mechanism. Therefore, even under similar linguistic inputs, agents may form diverging policy structures. As interactions proceed, the divergence $\Delta_{\text{intent}}$ tends to increase, leading to conflicting decisions or collaborative deadlocks.

This issue can be further exacerbated by attention decay in transformer-based models, wherein long-range dependencies—often encoding key intent signals—are progressively weakened. As observed in systems such as CAMEL~\cite{li2023camel}, such decay may contribute to agents drifting away from their initial goals and defaulting to irrelevant or linguistically plausible but semantically incoherent behaviors.

These observed misbehaviors are not incidental or correctable by surface-level prompt design. Rather, they stem from deep architectural mismatches between language model structures and the foundational demands of multi-agent collaboration. The lack of role-indexed semantic space, task-partitionable state space, and cross-agent policy synchronization renders current LLMs incapable of sustaining reliable collaboration~\cite{rashid2020monotonic,zhao2024graph}. This incompatibility must be addressed at the level of model design, not merely communication formatting.

\section{Toward a Native Multi-Agent Model Paradigm}
\subsection{Design Requirements}
Although current LLM-centric multi-agent systems demonstrate strong capabilities in language understanding and behavior generation, they exhibit structural limitations when confronted with collaborative tasks, long-term interactions, and systemic state evolution. As shown in the previous analysis, these issues stem not merely from the communication protocol itself, but from the lack of systematic modeling of structural requirements specific to multi-agent collaboration in the underlying model architecture. Therefore, constructing multi-agent systems with native collaborative capabilities necessitates incorporating the following fundamental architectural design requirements.

First, the model must incorporate a mechanism for \textbf{Role Persistence}, which ensures that each agent’s identity and behavioral boundaries are consistently maintained across multiple interaction rounds. In current language modeling paradigms, speaker roles are not explicitly encoded in token sequences and can only be inferred from prompts or context. This implicit modeling approach cannot ensure role consistency across turns. Hence, native multi-agent architectures should introduce a structured role space $R = \{r^{(1)}, \dots, r^{(n)}\}$, such that each agent’s state representation $h_t^{(i)}$ is explicitly bound to its role identifier:

\begin{equation}
\forall t, \quad h_t^{(i)} \sim p(h | r^{(i)}),
\end{equation}

This mechanism forms the foundation for maintaining identity coherence, task allocation, behavioral planning, and coordination in multi-agent systems.

Second, the model must support \textbf{Structured Communication}. In contrast to the linear token stream of natural language, inter-agent communication is better modeled as structured transmission over semantic graphs. Messages $m_{ij}^{(t)}$ should no longer be language fragments, but communication units encoding state, plans, and intentions in tensor form. Their purpose is not "interpretable dialogue" but "executable state exchange." Under this paradigm, communication is formalized as a transformation between agent states:

\begin{equation}
h_{t+1}^{(i)} = \mathcal{F}(h_t^{(i)}, m_{ji}^{(t)}), \quad m_{ji}^{(t)} = \mathcal{G}(h_t^{(j)}),
\end{equation}

where $\mathcal{G}$ is the message generation function and $\mathcal{F}$ is the parsing and state update function. The overall communication process should preserve structural homomorphism and semantic fidelity, such that $\mathcal{F} \circ \mathcal{G}$ approximates the identity mapping within the collaborative representation subspace.

Third, the model should enable \textbf{Inter-Agent State Synchronization}. In complex collaborative scenarios, agents must not only share current observations but also align on past strategies, intended goals, and behavioral constraints. This requires not only constructing explicit memory modules (e.g., multi-agent memory graphs $M_t = \{h_t^{(i)}\}_{i=1}^n$), but also establishing consistent alignment mechanisms that enforce:

\begin{equation}
\forall i, j, \quad \mathbb{E}_{s \sim \mathcal{S}} \left[ \| \phi(h_t^{(i)}; s) - \phi(h_t^{(j)}; s) \|^2 \right] \leq \epsilon,
\end{equation}

where $\phi(\cdot; s)$ denotes the semantic projection function under environment state $s$, and $\epsilon$ is an upper bound on alignment error. This synchronization capability is critical for resolving path conflicts, maintaining intention coherence, and ensuring task continuity in multi-agent systems.

Finally, native multi-agent systems should explicitly decouple three functional modules at the architectural level: \textbf{Perception-Action Loop}, \textbf{Planning}, and \textbf{Coordination Modeling}. These modules differ fundamentally in information flow, structural dependencies, and temporal scales:

\begin{itemize}
\item The perception-action module should support high-frequency, closed-loop feedback for real-time environmental responses.
\item The planning module should model long-horizon dependencies and trajectory evaluation.
\item The coordination module should be built on structured consistency, supporting hierarchical role modeling and interaction dependency graphs.
\end{itemize}

Current single-sequence language generation architectures often conflate these three components, resulting in behavioral entanglement, semantic path contamination, and distorted state evolution. Therefore, the evolution of multi-agent modeling paradigms should emphasize functional decomposition and structural decoupling, fostering modular and constraint-driven architectural design.

In summary, we argue that the future of multi-agent system development lies in native modeling paradigms that structurally satisfy the above four requirements. These mechanisms are not merely generalizations of language generation technologies, but structured responses to the intrinsic demands of multi-agent collaboration, forming the theoretical foundation for building stable, robust, and scalable agent systems.

\subsection{Modeling Implications}
Following the analysis of the structural limitations of natural language as a communication medium and the articulation of foundational design requirements for native multi-agent modeling, we further distill a set of systemic principles that should guide the modeling of language-based agents in collaborative multi-agent scenarios. These implications are not mere enhancements or fine-tuning strategies for existing models, but ontological reflections grounded in structural isomorphism and representational coupling regarding how large models should support multi-agent collaboration.

First, the \textbf{semantic space should enforce role-consistent structuring}. That is, the internal tensor representation of the model must distinguish between the semantic ownership of different agents, rather than encoding all tokens within a single unified vector manifold. This structural constraint necessitates the introduction of a semantic identity separation mechanism in the model—e.g., the construction of an agent-specific latent space with role-key embeddings—ensuring that state representations are stably bound to agent identities. Only by preserving role semantics at the representational level can the model support role persistence, task inheritance, and cross-turn state management.

Second, \textbf{communication should be modeled as a structured mapping function}, not merely as a by-product of language output. In the current LLM generation paradigm, an agent's output simultaneously carries behavioral intent and linguistic redundancy, which compromises the fidelity of state synchronization. Future models should treat communication as a learnable mapping between state tensors: $\mathcal{G}: \mathcal{H}^{(i)} \rightarrow \mathcal{M}_{ij}$, where $\mathcal{M}_{ij}$ denotes a structured message tensor sent from agent $i$ to agent $j$. This mechanism enhances communication efficiency and supports decoupled representation and gradient flow across messages.

Third, \textbf{action generation should not be conflated with language generation}. While language-based behavior planning may suffice in single-agent settings, multi-agent environments require action selection that accounts for inter-agent state dependencies, goal conflicts, and strategy coupling. Thus, models must decouple the execution channel from the language output logic, introducing an independent policy modeling path to avoid semantic contamination and behavioral interference between the two channels.

Fourth, \textbf{multi-agent models should maintain an explicit coordination graph}. Internally, the model should dynamically track a coordination state graph that encodes role relations, task dependencies, and semantic boundaries among agents. This structure can be formalized as a triplet graph:

\begin{equation}
\mathcal{G}_t = (\mathcal{A}, \mathcal{R}, \mathcal{E}_t), \quad \text{with } \mathcal{E}_t^{(i,j)} \sim \psi(h_t^{(i)}, h_t^{(j)}),
\end{equation}

where $\psi$ is the semantic relation modeling function and $\mathcal{E}_t$ denotes the time-varying coordination edges among agents. The existence of this graph enables the model to explicitly quantify semantic coupling strength, guiding communication frequency, content selection, and role scheduling strategies.

Finally, we emphasize the necessity of a \textbf{paradigm shift from pragmatics to structure}. Multi-agent interactions should no longer be regarded merely as an evolution of human-like dialogues, but as mechanisms for synchronization across representational spaces in complex systems. Communication, perception, planning, and behavior must be structurally reconstructed at the architectural level. Language is but one carrier, and its significance cannot replace the structural mechanisms required for systemic coordination. Just as deep learning progressed from surface classifiers to representational learning, multi-agent systems should evolve from natural language exchanges to alignment of internal representations.

In summary, building native multi-agent models requires advancing their structural adaptability in collaborative tasks by addressing five core perspectives: control of structural errors, modeling of role-state bindings, decoupling of language and behavior channels, tensorization of state communication, and graph-based coordination modeling. These modeling insights transcend specific applications and should form the theoretical foundation for a new paradigm in systemic multi-agent modeling.

\subsection{Open Challenges}
Although we have proposed a systematic set of structural requirements and modeling implications for advancing toward a native multi-agent modeling paradigm, significant challenges remain in building models with true structural coherence and generalized collaborative capabilities. Here, we outline several representative open problems as potential directions for future research.

While this paper emphasizes the structural limitations of natural language in achieving state alignment and semantic synchronization among agents, natural language retains irreplaceable advantages in human-system interaction due to its interpretability and debuggability. In system development, behavior monitoring, and policy evolution, token-level natural language interaction provides an intuitive, low-barrier diagnostic channel for researchers and developers. A key open question is: How can we retain natural language as a human-facing debugging interface while designing structured communication protocols and internal semantic tensor alignment mechanisms? Addressing this requires controllable and invertible mappings from internal semantic states to natural language explanations (e.g., structured explanation generation), along with interface isolation mechanisms that prevent debugging pathways from contaminating multi-agent policies.

In real-world multi-agent systems, modalities such as language, vision, spatial reasoning, and dynamics must be jointly modeled and shared. However, current LLM-based multi-agent architectures are predominantly language-centric, with other modalities integrated via external perception modules. This design results in temporal structure mismatches, sampling frequency disparities, and semantic dimensionality misalignments across modalities, hindering state sharing and intention coordination. The core modeling challenge is to construct a unified multimodal state representation space, where language generation reflects manipulable distributions over perceptual and intentional tensors—not merely behavior descriptions. To this end, future multi-agent systems should evolve beyond loose dialog frameworks toward brain-inspired collaborative architectures: each agent functions as a specialized cortical region, achieving global state coupling via high-speed, low-redundancy, and structurally aligned semantic channels. Such systems would no longer rely on token-level interaction, but instead achieve synchronization through cascaded high-dimensional state activations. In the human brain, communication among vision, motor, and language areas is not mediated by discrete symbols, but via tightly coupled neural states that support real-time alignment and functional specialization. Implementing this in multi-agent AI demands neural-style semantic tensor sharing structures with modality-invariance and channel decoupling, to support semantic consistency and intention linkage across modalities and agents. This poses a fundamental architectural challenge, especially in designing unified latent spaces. Potential directions include building structurally aligned multimodal encoder-decoder models or developing integrated tensor spaces across perception, intention, and action based on heterogeneous information graphs, enabling a neural-like coordination paradigm.

In addition, real-world multi-agent tasks often involve dynamically expanding agent populations, role topologies, and task workflows, dramatically increasing the dimensionality of the modeling space. This raises the problem of structural generalization: Can a native multi-agent model generalize to unseen coordination structures in a zero-shot manner? Traditional language models excel at sequence-level generalization but struggle to transfer over structural variations. Enabling agent-to-agent generalization requires compositional generalization over graph structures, allowing adaptation to novel role compositions, coordination dependencies, and communication protocols. Addressing this challenge demands new structure-aware objectives, general coordination graph representations, and dedicated benchmarks for structural generalization. Moreover, structural generalization should not be limited to static graph-level configurations but extended to dynamic interaction trajectories. Real-world multi-agent collaboration unfolds over multiple rounds of asynchronous and context-sensitive exchanges. Thus, an effective model must generalize not only over task graphs but also over multi-turn interaction structures, adapting to evolving coordination demands. This requires the model to maintain a global view of interaction history and selectively activate relevant sub-networks or policy pathways based on current dialogue states, akin to the task-driven activation of specialized cortical regions in the human brain. Unlike the monolithic input-output pipeline common in current LLM architectures, future models should embody interaction-aware computation paths, where the execution flow is conditioned on communication states and historical semantics.

As agent communication frequency and state coupling increase, the system's security boundary and robustness become more fragile. A single communication error can cascade through state transition chains, triggering policy divergence. Most current multi-agent systems rely on soft, token-level interactions and lack design principles for adversarial robustness, coordination failure recovery, or communication noise isolation. Thus, a core modeling problem is: How can we introduce architectural mechanisms such as redundant role mappings, mutual plan checking, and error rollback to build fault-tolerant coordination models?

Finally, native multi-agent systems introduce a new structure of decision fluidity, where agent behaviors are dynamically dependent on others' states and communications. This blurs responsibility boundaries: Is a failed action due to perception bias, misleading messages, or unclear goal specifications? This introduces unprecedented challenges in system verifiability and accountability. Especially in real-world deployment, it is essential to enable causal attribution mechanisms based on communication graphs and state trajectories, supporting responsibility tracing and interpretable outputs—preconditions for compliance and trust.

These open challenges are not isolated but interconnected across modeling, system architecture, interaction design, security, and governance. We argue that constructing deployable, verifiable, and truly collaborative native multi-agent systems requires overcoming not only model-level limitations, but also establishing cross-layered consensus spanning paradigms, system structure, and theoretical foundations. The resolution of these issues will determine whether multi-agent AI systems can evolve from "language-driven dialog simulations" to "structure-aware collaborative intelligence."

\section{Conclusion and Outlook}



This paper presents a systematic reflection on the prevailing design paradigm in multi-agent systems, in which natural language serves as the primary communication protocol. While natural language offers strong interpretability and debuggability in human–machine interaction, we argue that it is structurally misaligned with the high-dimensional semantic spaces in which LLMs internally operate. This semantic misalignment leads to irreversible information compression and intention drift during inter-agent communication, and further induces systemic issues such as role inconsistency, behavioral decoupling, and collaboration collapse.

Beyond communication-level limitations, we contend that current LLM training paradigms are fundamentally ill-suited for collaborative settings characterized by multi-role coordination, task dependency, and state synchronization. The standard single-stream, uni-perspective generation mechanism of LLMs inherently lacks support for semantic binding, persistent role tracking, and behavioral consistency—requirements that are critical for scalable and reliable multi-agent collaboration. This limitation reflects a deeper architectural incompatibility between model design and interactive multi-agent tasks.

In response, we advocate for a native multi-agent modeling paradigm, one that shifts away from simulating human-like dialogues toward satisfying the structural demands of semantic alignment and coordination fidelity. We propose four key architectural components essential for this paradigm: role persistence modeling, structured communication mechanisms, inter-agent state synchronization, and functional decoupling across perception, planning, and coordination. We further highlight that current LLMs lack intrinsic support for these dimensions.

Looking forward, several critical challenges must be addressed. First, developing hybrid communication channels that integrate both structured tensor-level semantics and natural language for human interpretability is critical for enabling both human–agent alignment and autonomous system behavior. Second, the construction of unified representational spaces capable of modeling multi-modal, multi-role, and multi-task dynamics remains a fundamental challenge for semantic invariance and behavioral coherence. Third, as agent interactions grow in complexity, robustness, verifiability, and causal attribution should be explicitly incorporated into model design and evaluation pipelines, especially in open-world deployment scenarios.

We call for a fundamental rethinking of the nature of “communication” in multi-agent systems. Rather than relying on natural language as a generalized proxy, agent communication should be redefined as the structured transmission of internal state and coordination information. We hope this work lays a theoretical foundation for the development of structurally aligned multi-agent systems and encourages further interdisciplinary research into native paradigms for collaborative intelligence.









\bibliographystyle{IEEEtran}
\bibliography{SEAMS}  

\begin{thebibliography}{10}
\providecommand{\url}[1]{#1}
\csname url@samestyle\endcsname
\providecommand{\newblock}{\relax}
\providecommand{\bibinfo}[2]{#2}
\providecommand{\BIBentrySTDinterwordspacing}{\spaceskip=0pt\relax}
\providecommand{\BIBentryALTinterwordstretchfactor}{4}
\providecommand{\BIBentryALTinterwordspacing}{\spaceskip=\fontdimen2\font plus
\BIBentryALTinterwordstretchfactor\fontdimen3\font minus \fontdimen4\font\relax}
\providecommand{\BIBforeignlanguage}[2]{{%
\expandafter\ifx\csname l@#1\endcsname\relax
\typeout{** WARNING: IEEEtran.bst: No hyphenation pattern has been}%
\typeout{** loaded for the language `#1'. Using the pattern for}%
\typeout{** the default language instead.}%
\else
\language=\csname l@#1\endcsname
\fi
#2}}
\providecommand{\BIBdecl}{\relax}
\BIBdecl

\bibitem{wasif2025multi}
M.~Wasif and D.~Tunkel, ``Multi-agent collaboration in ai: Enhancing software development with autonomous llms,'' 2025.

\bibitem{financialtimes2025aiagents}
{Financial Times}, ``Ai agents: from co-pilot to autopilot,'' \url{https://www.ft.com/content/3e862e23-6e2c-4670-a68c-e204379fe01f}, 2025, accessed: 2025-05-28.

\bibitem{li2025large}
P.~Li, Z.~An, S.~Abrar, and L.~Zhou, ``Large language models for multi-robot systems: A survey,'' \emph{arXiv preprint arXiv:2502.03814}, 2025.

\bibitem{relevanceai2024mas}
\BIBentryALTinterwordspacing
{Relevance AI}, ``What is a multi-agent system,'' 2024, accessed: 2025-05-28. [Online]. Available: \url{https://relevanceai.com/learn/what-is-a-multi-agent-system}
\BIBentrySTDinterwordspacing

\bibitem{wikipedia_multiagent}
{Wikipedia contributors}, ``Multi-agent system --- wikipedia{,} the free encyclopedia,'' \url{https://en.wikipedia.org/wiki/Multi-agent_system}, 2025, accessed: 2025-05-28.

\bibitem{lyu2025llmcoop}
X.~Lyu, ``Llms for multi-agent cooperation,'' \url{https://xue-guang.com/post/llm-marl/}, 2025, accessed: 2025-05-28.

\bibitem{ehtesham2025survey}
A.~Ehtesham, A.~Singh, G.~K. Gupta, and S.~Kumar, ``A survey of agent interoperability protocols: Model context protocol (mcp), agent communication protocol (acp), agent-to-agent protocol (a2a), and agent network protocol (anp),'' \emph{arXiv preprint arXiv:2505.02279}, 2025.

\bibitem{google2025a2a}
Google, ``Announcing the agent2agent protocol (a2a),'' \url{https://developers.googleblog.com/en/a2a-a-new-era-of-agent-interoperability/}, 2025, accessed: 2025-05-29.

\bibitem{anthropic2024mcp}
Anthropic, ``Introducing the model context protocol,'' \url{https://www.anthropic.com/news/model-context-protocol}, 2024, accessed: 2025-05-29.

\bibitem{yi2024survey}
Z.~Yi, J.~Ouyang, Y.~Liu, T.~Liao, Z.~Xu, and Y.~Shen, ``A survey on recent advances in llm-based multi-turn dialogue systems,'' \emph{arXiv preprint arXiv:2402.18013}, 2024.

\bibitem{autogen2025}
Microsoft, ``Multi-agent conversation framework: Autogen 0.2 documentation,'' \url{https://microsoft.github.io/autogen/0.2/docs/Use-Cases/agent_chat/}, 2025, accessed: 2025-05-28.

\bibitem{zhou2025continuum}
P.~Zhou, Z.~Nie, and H.~Li, ``Continuum-interaction-driven intelligence: Human-aligned neural architecture via crystallized reasoning and fluid generation,'' \emph{arXiv preprint arXiv:2504.09301}, 2025.

\bibitem{guan2025evaluating}
S.~Guan, H.~Xiong, J.~Wang, J.~Bian, B.~Zhu, and J.-g. Lou, ``Evaluating llm-based agents for multi-turn conversations: A survey,'' \emph{arXiv preprint arXiv:2503.22458}, 2025.

\bibitem{laban2025llms}
P.~Laban, H.~Hayashi, Y.~Zhou, and J.~Neville, ``Llms get lost in multi-turn conversation,'' \emph{arXiv preprint arXiv:2505.06120}, 2025.

\bibitem{han2024llm}
S.~Han, Q.~Zhang, Y.~Yao, W.~Jin, Z.~Xu, and C.~He, ``Llm multi-agent systems: Challenges and open problems,'' \emph{arXiv preprint arXiv:2402.03578}, 2024.

\bibitem{pungas2023autogpt}
T.~Pungas, ``Why autogpt fails and how to fix it,'' \url{https://www.taivo.ai/__why-autogpt-fails-and-how-to-fix-it/}, 2023, accessed: 2025-05-28.

\bibitem{agentverse2024}
{SmythOS}, ``Agentverse vs. autogpt: Comparing ai agent development platforms,'' \url{https://smythos.com/ai-agents/comparison/agentverse-vs-autogpt/}, 2024, accessed: 2025-05-28.

\bibitem{cemri2025multi}
M.~Cemri, M.~Z. Pan, S.~Yang, L.~A. Agrawal, B.~Chopra, R.~Tiwari, K.~Keutzer, A.~Parameswaran, D.~Klein, K.~Ramchandran \emph{et~al.}, ``Why do multi-agent llm systems fail?'' \emph{arXiv preprint arXiv:2503.13657}, 2025.

\bibitem{zhang2025agent}
S.~Zhang, M.~Yin, J.~Zhang, J.~Liu, Z.~Han, J.~Zhang, B.~Li, C.~Wang, H.~Wang, Y.~Chen \emph{et~al.}, ``Which agent causes task failures and when? on automated failure attribution of llm multi-agent systems,'' \emph{arXiv preprint arXiv:2505.00212}, 2025.

\bibitem{turpin2023language}
M.~Turpin, J.~Michael, E.~Perez, and S.~Bowman, ``Language models don't always say what they think: Unfaithful explanations in chain-of-thought prompting,'' \emph{Advances in Neural Information Processing Systems}, vol.~36, pp. 74\,952--74\,965, 2023.

\bibitem{chen2025reasoning}
Y.~Chen, J.~Benton, A.~Radhakrishnan, J.~Uesato, C.~Denison, J.~Schulman, A.~Somani, P.~Hase, M.~Wagner, F.~Roger \emph{et~al.}, ``Reasoning models don't always say what they think,'' \emph{arXiv preprint arXiv:2505.05410}, 2025.

\bibitem{wang2023fake}
Y.~Wang, Y.~Teng, K.~Huang, C.~Lyu, S.~Zhang, W.~Zhang, X.~Ma, Y.-G. Jiang, Y.~Qiao, and Y.~Wang, ``Fake alignment: Are llms really aligned well?'' \emph{arXiv preprint arXiv:2311.05915}, 2023.

\bibitem{zhang2024cut}
G.~Zhang, Y.~Yue, Z.~Li, S.~Yun, G.~Wan, K.~Wang, D.~Cheng, J.~X. Yu, and T.~Chen, ``Cut the crap: An economical communication pipeline for llm-based multi-agent systems,'' \emph{arXiv preprint arXiv:2410.02506}, 2024.

\bibitem{galileo2025multiagent}
G.~AI, ``Multi-agent ai: Performance metrics \& evaluation frameworks,'' \url{https://galileo.ai/blog/success-multi-agent-ai}, 2025, accessed: 2025-05-28.

\bibitem{mordatch2018emergence}
I.~Mordatch and P.~Abbeel, ``Emergence of grounded compositional language in multi-agent populations,'' in \emph{Proceedings of the AAAI conference on artificial intelligence}, vol.~32, no.~1, 2018.

\bibitem{futrell2022information}
R.~Futrell and M.~Hahn, ``Information theory as a bridge between language function and language form,'' \emph{Frontiers in Communication}, vol.~7, p. 657725, 2022.

\bibitem{lazaridou2020multi}
A.~Lazaridou, A.~Potapenko, and O.~Tieleman, ``Multi-agent communication meets natural language: Synergies between functional and structural language learning,'' in \emph{Proceedings of the 58th Annual Meeting of the Association for Computational Linguistics}, 2020, pp. 7663--7674.

\bibitem{hu2025position}
J.~Hu, Y.~Dong, S.~Ao, Z.~Li, B.~Wang, L.~Singh, G.~Cheng, S.~D. Ramchurn, and X.~Huang, ``Position: Towards a responsible llm-empowered multi-agent systems,'' \emph{arXiv preprint arXiv:2502.01714}, 2025.

\bibitem{yan2025beyond}
B.~Yan, X.~Zhang, L.~Zhang, L.~Zhang, Z.~Zhou, D.~Miao, and C.~Li, ``Beyond self-talk: A communication-centric survey of llm-based multi-agent systems,'' \emph{arXiv preprint arXiv:2502.14321}, 2025.

\bibitem{li2023camel}
G.~Li, H.~Hammoud, H.~Itani, D.~Khizbullin, and B.~Ghanem, ``Camel: Communicative agents for" mind" exploration of large language model society,'' \emph{Advances in Neural Information Processing Systems}, vol.~36, pp. 51\,991--52\,008, 2023.

\bibitem{crewai2024}
J.~Moura and contributors, ``Crewai: A multi-agent orchestration framework for llm agents,'' \url{https://github.com/joaomdmoura/crewAI}, 2024, gitHub Repository.

\bibitem{xiao2024tradingagents}
Y.~Xiao, E.~Sun, D.~Luo, and W.~Wang, ``Tradingagents: Multi-agents llm financial trading framework,'' \emph{arXiv preprint arXiv:2412.20138}, 2024.

\bibitem{kampman2024ai}
O.~P. Kampman, Y.~S. Phang, S.~Han, M.~Xing, X.~Hong, H.~Hoosainsah, C.~Tan, G.~I. Winata, S.~Wang, C.~Heaukulani \emph{et~al.}, ``An ai-assisted multi-agent dual dialogue system to support mental health care providers,'' \emph{arXiv preprint arXiv:2411.18429}, 2024.

\bibitem{zhang2024simulating}
Z.~Zhang, D.~Zhang-Li, J.~Yu, L.~Gong, J.~Zhou, Z.~Hao, J.~Jiang, J.~Cao, H.~Liu, Z.~Liu \emph{et~al.}, ``Simulating classroom education with llm-empowered agents,'' \emph{arXiv preprint arXiv:2406.19226}, 2024.

\bibitem{wei2022chain}
J.~Wei, X.~Wang, D.~Schuurmans, M.~Bosma, F.~Xia, E.~Chi, Q.~V. Le, D.~Zhou \emph{et~al.}, ``Chain-of-thought prompting elicits reasoning in large language models,'' \emph{Advances in neural information processing systems}, vol.~35, pp. 24\,824--24\,837, 2022.

\bibitem{mialon2023augmented}
G.~Mialon, R.~Dess{\`\i}, M.~Lomeli, C.~Nalmpantis, R.~Pasunuru, R.~Raileanu, B.~Rozi{\`e}re, T.~Schick, J.~Dwivedi-Yu, A.~Celikyilmaz \emph{et~al.}, ``Augmented language models: a survey,'' \emph{arXiv preprint arXiv:2302.07842}, 2023.

\bibitem{shinn2023reflexion}
N.~Shinn, F.~Cassano, A.~Gopinath, K.~Narasimhan, and S.~Yao, ``Reflexion: Language agents with verbal reinforcement learning,'' \emph{Advances in Neural Information Processing Systems}, vol.~36, pp. 8634--8652, 2023.

\bibitem{finin1994kqml}
T.~Finin, R.~Fritzson, D.~McKay, and R.~McEntire, ``Kqml as an agent communication language,'' in \emph{Proceedings of the third international conference on Information and knowledge management}, 1994, pp. 456--463.

\bibitem{fipa2000fipa}
T.~FIPA, ``Fipa communicative act library specification,'' \emph{Change}, vol. 2000, no. 01/18, 2000.

\bibitem{wooldridge2009introduction}
M.~Wooldridge, \emph{An introduction to multiagent systems}.\hskip 1em plus 0.5em minus 0.4em\relax John wiley \& sons, 2009.

\bibitem{foerster2016learning}
J.~Foerster, I.~A. Assael, N.~De~Freitas, and S.~Whiteson, ``Learning to communicate with deep multi-agent reinforcement learning,'' \emph{Advances in neural information processing systems}, vol.~29, 2016.

\bibitem{lazaridou2020emergent}
A.~Lazaridou and M.~Baroni, ``Emergent multi-agent communication in the deep learning era,'' \emph{arXiv preprint arXiv:2006.02419}, 2020.

\bibitem{lowe2019pitfalls}
R.~Lowe, J.~Foerster, Y.-L. Boureau, J.~Pineau, and Y.~Dauphin, ``On the pitfalls of measuring emergent communication,'' in \emph{Proceedings of the 18th International Conference on Autonomous Agents and MultiAgent Systems}, 2019, pp. 693--701.

\bibitem{kottur2017natural}
S.~Kottur, J.~Moura, S.~Lee, and D.~Batra, ``Natural language does not emerge ‘naturally’in multi-agent dialog,'' in \emph{Proceedings of the 2017 Conference on Empirical Methods in Natural Language Processing}, 2017, pp. 2962--2967.

\bibitem{jacob2021multitasking}
A.~P. Jacob, M.~Lewis, and J.~Andreas, ``Multitasking inhibits semantic drift,'' in \emph{Proceedings of the 2021 Conference of the North American Chapter of the Association for Computational Linguistics: Human Language Technologies}, 2021, pp. 5351--5366.

\bibitem{smythos2025comparing}
{SmythOS}, ``Comparing agent communication languages and protocols: Choosing the right framework for multi-agent systems,'' \url{https://smythos.com/developers/ai-agent-development/agent-communication-languages-and-protocols-comparison/}, 2025.

\bibitem{piatti2024cooperate}
G.~Piatti, Z.~Jin, M.~Kleiman-Weiner, B.~Sch{\"o}lkopf, M.~Sachan, and R.~Mihalcea, ``Cooperate or collapse: Emergence of sustainable cooperation in a society of llm agents,'' \emph{Advances in Neural Information Processing Systems}, vol.~37, pp. 111\,715--111\,759, 2024.

\bibitem{zou2025survey}
H.~P. Zou, W.-C. Huang, Y.~Wu, Y.~Chen, C.~Miao, H.~Nguyen, Y.~Zhou, W.~Zhang, L.~Fang, L.~He \emph{et~al.}, ``A survey on large language model based human-agent systems,'' \emph{arXiv preprint arXiv:2505.00753}, 2025.

\bibitem{brown2017multiagent}
P.~N. Brown, H.~P. Borowski, and J.~R. Marden, ``Are multiagent systems resilient to communication failures?'' \emph{arXiv preprint arXiv:1710.08500}, 2017.

\bibitem{sun2025idiosyncrasies}
M.~Sun, Y.~Yin, Z.~Xu, J.~Z. Kolter, and Z.~Liu, ``Idiosyncrasies in large language models,'' \emph{arXiv preprint arXiv:2502.12150}, 2025.

\bibitem{birr2024autogpt+}
T.~Birr, C.~Pohl, A.~Younes, and T.~Asfour, ``Autogpt+ p: Affordance-based task planning with large language models,'' \emph{arXiv preprint arXiv:2402.10778}, 2024.

\bibitem{harnad1990symbol}
S.~Harnad, ``The symbol grounding problem,'' \emph{Physica D: Nonlinear Phenomena}, vol.~42, no. 1-3, pp. 335--346, 1990.

\bibitem{baroni2020linguistic}
M.~Baroni, ``Linguistic generalization and compositionality in modern artificial neural networks,'' \emph{Philosophical Transactions of the Royal Society B}, vol. 375, no. 1791, p. 20190307, 2020.

\bibitem{zhang2024language}
M.~Zhang, O.~Press, W.~Merrill, A.~Liu, and N.~A. Smith, ``How language model hallucinations can snowball,'' in \emph{International Conference on Machine Learning}.\hskip 1em plus 0.5em minus 0.4em\relax PMLR, 2024, pp. 59\,670--59\,684.

\bibitem{agenticAIguide}
{VideoSDK Team}, ``Agentic ai: A complete guide to autonomous ai agents,'' \url{https://www.videosdk.live/developer-hub/ai_agent/agentic-ai}, 2024, accessed: 2025-06-01.

\bibitem{autogpt2023issue5190}
{Significant Gravitas Contributors}, ``Auto-gpt performance: Issue \#5190,'' \url{https://github.com/Significant-Gravitas/AutoGPT/issues/5190}, 2023, gitHub Issue; Accessed: 2025-06-01.

\bibitem{rashid2020monotonic}
T.~Rashid, M.~Samvelyan, C.~S. De~Witt, G.~Farquhar, J.~Foerster, and S.~Whiteson, ``Monotonic value function factorisation for deep multi-agent reinforcement learning,'' \emph{Journal of Machine Learning Research}, vol.~21, no. 178, pp. 1--51, 2020.

\bibitem{zhao2024graph}
B.~Zhao, M.~Huo, Z.~Li, Z.~Yu, and N.~Qi, ``Graph-based multi-agent reinforcement learning for large-scale uavs swarm system control,'' \emph{Aerospace Science and Technology}, vol. 150, p. 109166, 2024.

\end{thebibliography}

\end{document}